\documentclass[final,3p,times,twocolumn,authoryear]{elsarticle}

\usepackage{natbib}
\setcitestyle{numbers,square}

\usepackage{amsmath,amssymb,amsfonts}
\usepackage{algorithmic}
\usepackage{graphicx}
\usepackage{algorithm,algorithmic}
\usepackage{hyperref}
\hypersetup{hypertex=true,
            colorlinks=true,
            linkcolor=blue,
            anchorcolor=blue,
            citecolor=blue}
\usepackage{textcomp}

\usepackage[utf8]{inputenc}
\usepackage[nameinlink]{cleveref}
\usepackage{latexsym}
\usepackage{url}
\usepackage{xcolor}
\usepackage[marginal]{footmisc}

\crefname{figure}{\textcolor{blue}{Fig.}}{\textcolor{blue}{Fig.}}
\crefname{sec}{\textcolor{blue}{Section }}{\textcolor{blue}{Section.}}

\usepackage{epstopdf}
\usepackage{color}
\usepackage{gensymb}
\usepackage{subfig}
\usepackage{tabularx,ragged2e,booktabs,caption}
\usepackage{textcomp}
\usepackage{ragged2e}\usepackage{booktabs}\usepackage{caption}
\usepackage{colortbl}
\usepackage{tabu}
\usepackage{multirow}
\usepackage{pifont}
\usepackage{lscape}
\usepackage{makecell}
\usepackage{etoolbox}
\usepackage{amssymb}

\begin{document}

\begin{frontmatter}



\title{Evidential learning driven Breast Tumor Segmentation with Stage-divided Vision-Language Interaction}


\author{Jingxing Zhong$^{a,b,1}$, Qingtao Pan$^{c,1}$, Xuchang Zhou$^{d,1}$, Jiazhen Lin$^{e,*}$, Xinguo Zhuang$^{a,*}$}

\affiliation[$a$]{organization={Center for Precision Medicine, The First Affiliated Hospital of Xiamen University, School of Medicine, Xiamen University, Xiamen, China; Department of Laboratory Medicine, The First Affiliated Hospital of Xiamen University, School of Medicine, Xiamen University},
            city={Xiamen},
            postcode={361003},
            country={China}}

\affiliation[$b$]{organization={Maynooth International Engineering College, Fuzhou University},
            city={Fuzhou},
            postcode={350108}, 
            country={China}}

\affiliation[$c$]{organization={Department of Computer and Data Science, Case Western Reserve University, Cleveland, Ohio, USA},
            city={Cleveland},
            postcode={44106},
			state={Ohio},
            country={USA}}

\affiliation[$d$]{organization={Department of Rehabilitation Medicine, the First Affiliated Hospital of Xiamen University, School of Medicine, Xiamen University},
            city={Xiamen},
            postcode={361003},
            country={China}}

\affiliation[$e$]{organization={Department of Pharmacy, Xiang'an Hospital of Xiamen University, School of Medicine, Xiamen University},
            city={Xiamen},
            postcode={361005}, 
            country={China}}

\cortext[cor1]{Corresponding authors. \\ E-mail addresses: jiazhen\_lin@163.com (J. Lin) \\ zxg579@126.com (X. Zhuang) \\
1 denotes these authors contributed equally to this work.}

\begin{abstract}
Breast cancer is one of the most common causes of death among women worldwide, with millions of fatalities annually. Magnetic Resonance Imaging (MRI) can provide various sequences for characterizing tumor morphology and internal patterns, and becomes an effective tool for detection and diagnosis of breast tumors. However, previous deep-learning based tumor segmentation methods have limitations in accurately locating tumor contours due to the challenge of low contrast between cancer and normal areas and blurred boundaries. Leveraging text prompt information holds promise in ameliorating tumor segmentation effect by delineating segmentation regions. Inspired by this, we propose text-guided Breast Tumor Segmentation model (TextBCS) with stage-divided vision-language interaction and evidential learning. Specifically, the proposed stage-divided vision-language interaction facilitates information mutual between visual and text features at each stage of down-sampling, further exerting the advantages of text prompts to assist in locating lesion areas in low contrast scenarios. Moreover, the evidential learning is adopted to quantify the segmentation uncertainty of the model for blurred boundary. It utilizes the variational Dirichlet to characterize the distribution of the segmentation probabilities, addressing the segmentation uncertainties of the boundaries. Extensive experiments validate the superiority of our TextBCS over other segmentation networks, showcasing the best breast tumor segmentation performance on publicly available datasets.
\end{abstract}


\end{frontmatter}


\begin{figure}[h]
\centering
\includegraphics[width=\linewidth]{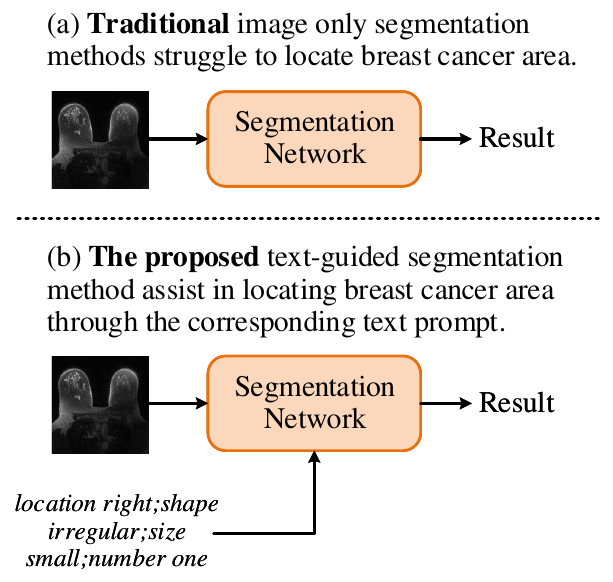}
\caption{Various segmentation methods for breast cancer segmentation. (a) Traditional segmentation methods rely solely on the image to perform segmentation tasks. (b) Our TextBCS leverages text prompts related to canceration regions to assist in locating breast cancer areas.}
\label{fig1}
\end{figure}

\section{Introduction}
\label{sec1}
Breast cancer has been the second most usual cause of cancer death among women \cite{ref1,ref2}, which is a type of cancer that develops in the cells of the breast. According to data from 2021 Cancer Statistics, breast cancer has surpassed lung cancer and ranks first among all new female cancers \cite{ref3}. Fortunately, the existing literatures reveal that early identification of malignant tumors with timely clinical intervention will considerably improve patient survival and outcomes \cite{ref4}. Currently, there are a series of image methods for breast cancer screening and monitoring \cite{ref5, ref6}, such as computed tomography (CT), mammography, positron emission tomography (PET), and magnetic resonance imaging (MRI). Among these methods, MRI is a commonly used imaging tool for breast screening because it is a highly sensitive modality for detecting breast cancer with reported sensitivity of more than 80\%. However, MRI is a conventional structure imaging, which is acquired at high spatial resolution without temporal resolution. Recently, Dynamic contrast-enhanced magnetic resonance image (DCE-MRI) utilizes the injection of contrast agent for unparalleled tissue contrast. This unique feature differentiates DCE-MRI from other MRI sequences, enabling more pronounced intensity contrast between tumor and background. 

While DCE-MRI provides high sensitivity due to dynamic contrast enhancement, it can still exhibit low local contrast between tumor regions and adjacent normal tissues \cite{ref62}. This subtle intensity difference poses a challenge for breast tumor segmentation. In recent years, deeping learning has shown an enormous impact on medical image analysis, including registration \cite{ref7}, segmentation \cite{ref8}, and diagnosis \cite{ref9}. Those techniques can also been applied for precise breast cancer segmentation. Hence, designing a powerful breast cancer segmentation model is a crucial procedure for promoting the development of breast cancer diagnosis.

Recently, several deep learning-based segmentation methods \cite{ref10,ref11,ref12,ref13,ref14,ref15} are proposed for breast cancer segmentation (as shown in Fig. \ref{fig1}(a)), which contains two major categories, i.e., network architecture and optimization protocols. For network architecture, a multi-stream fusion network \cite{ref11} was proposed to selectively fuse the useful information from different modalities and suppress the respective noise signals. Subsequently, a multi-scale context extractor block \cite{ref14} consisting of atrous convolutions with different sampling rates was introduced to extract multi-scale image features. Additionally, in \cite{ref15}, a novel 3D aﬃnity learning based multi-branch ensemble network for accurate breast tumor segmentation. The performance of these network architectures is indeed superior to lightweight networks. For optimization protocols, Zhang et al \cite{ref10} proposed a Dice-Sensitivity like loss function and a reinforcement sampling strategy to handle the class-imbalance problem. To suppress false positive segmentation with similar contrast enhancement characteristics to true breast tumors, tumor-sensitive synthesis module \cite{ref13} was proposed to feedback differential loss of the true and false breast tumors. Although these two types of methods have achieved high segmentation performance, learning solely from the image modality remains challenging for cases with low contrast and blurred boundaries.

To solve these challenges, we propose a Text-guided Breast Tumor Segmentation model (TextBCS, as shown in Fig. \ref{fig1}(b)), which contains a Stage-divided Vision-Language Interaction module (SVLI) and the Evidential Learning (EL) strategy. The SVLI is proposed to enhance information mutual between visual and text features at each stage of down-sampling, assisting in locating lesion areas in low contrast scenarios. The EL is designed to quantify the segmentation uncertainty of the model for blurred boundaries. In EL, the variational Dirichlet parameterized with collected evidence is introduced to characterize the distribution of the result probabilities, reducing the segmentation uncertainties of boundaries.

Our contributions are summarized as follows:

\begin{itemize}

    \item This is the first text-guided breast tumor segmentation method (TextBCS) in DCE-MRI. It enables the model to focuse more specifically on areas related to breast cancer in the image through utilizing the knowledge compensation of the text prompt.

    \item A stage-divided vision-language interaction module is proposed to smoothly achieves the fusion of visual-language information. It enables each level of image features to receive guidance from text prompts, enhancing the model's ability to recognize lesions.
    
    \item The evidential learning is incorporated to explicitly estimate the uncertainty of segmentation probabilities. It introduces Dirichlet distribution to model the degree of belief for each category, driving the model to maintain high uncertainty in the case of blurred boundaries, thus avoiding being misled by "false signals".
        
    \item Experiment results on a public breast cancer segmentation dataset demonstrate the effectiveness of each component of our method and the superiority of our TextBCS over the state-of-the-art methods.
\end{itemize}

\section{Related works}
\label{sec2}

\subsection{Breast Cancer Segmentation}
Tumor segmentation is an important method in the diagnosis of breast cancer \cite{ref16}. Recently, CNN-based methods have dominated the field of automatic breast cancer segmentation, including ultrasound \cite{ref17,ref18,ref19}, mammography \cite{ref20}, pathology \cite{ref21}, and DCE-MRI \cite{ref22,ref23}. Zhao et al. \cite{ref17} proposed a novel pyramidal pooling U-Net network (PPU-Net) to segment breast lesion by extracting more scale information. Lei et al. \cite{ref18} developed a deep learning encoder-decoder segmentation method based on a self-co-attention mechanism. It designed a non-local context block (NCB) to augment the learning of high-level contextual cues. Wang et al. \cite{ref19} proposed a fusion network including an encoder path, a decoder path, and a core fusion stream path (FSP). It also designed a weighted-balanced loss function to address the problem of lesion regions having different sizes. In addition, Li et al. \cite{ref20} introduced a novel end-to-end deep learning framework for mammogram image processing, which computes mass segmentation and simultaneously predicts diagnosis results. Carlos et al. \cite{ref21} introduced an end-to-end multi-task framework designed to leverage the inherent correlations between breast cancer lesion classification and segmentation tasks. Furthermore, MRI has become a crucial diagnostic tool for breast cancer owing to its excellent sensitivity and specificity in tumor detection. For breast tumor segmentation in MRI, researchers have utilized both single-modality and multi-modality segmentation approaches. For example, Zhou et al. \cite{ref15} built two different types of subnetworks to form a multi-branch network and proposed an end-to-end trainable 3D affinity learning based refinement module to discover more pixels belonging to breast tumors. In \cite{ref25}, two parallel convolutional networks are designed to retrieve the maps of apparent diffusion coefficient (ADC) and T2W images, which are then integrated to use the complimentary information in MP-MRI. By utilizing extrusion and excitation blocks, it is feasible to automatically increase the number of relevant features in the fusion feature map.

\subsection{Text-Guided Learning}
In medical image segmentation, particularly for lesion segmentation, integrating text-guided learning oﬀers a novel solution to address the limitations of traditional image-only methods. In the field of natural image segmentation, VTL \cite{ref26} and LAVT \cite{ref27} are classical methods. Additionally, UniLSeg \cite{ref28} combines image and text to improve adaptability across tasks, performing well with multi-task data and automatic annotation. Recently, CMIRNet \cite{ref29} devised a Text-Guided Multi-Modality Joint Encoder (TGMM-JE) to extract key expression and encoded the important visual features under the continuous guidance of language expression. It also designed a Cross-Graph Interactive Positioning (CGIP) module to locate the key pixels of the referent object in deepest layer. Nevertheless, the complexity and ambiguity of medical images pose substantial challenges for these rigid alignment methods. As a result, customized approaches have been developed to meet the specific demands of medical imaging, integrating text information at different stages of the encoder-decoder framework. For instance, models like CLIP \cite{ref30}, GLoRIA \cite{ref31}, ConVIRT \cite{ref32} enable image encoders to learn textual information for downstream segmentation tasks, while ConTEXTual Net \cite{ref33} and TGANet \cite{ref34} incorporate text in the decoder for image segmentation. Additionally, LViT \cite{ref35} integrates text during each down-sampling of the encoder to enhance the advantage of text prompt, while C2FVL \cite{ref36} and VLSM \cite{ref37} introduce text features at the skip connections. In addition, some works \cite{h2,h3} utilize text prompts to improve semi supervised segmentation. However, the above methods only rely on the shallow interaction between text and images, which restricts the effective alignment and fusion between image and text. 

\subsection{Evidential Learning}
In recent years, several uncertainty quantification approaches leveraging the Evidential Learning (EL) have been introduced to the field of computer vision. Unlike traditional probabilistic methods such as Softmax, evidence learning not only outputs category probabilities, but also prediction uncertainties through buliding Dirichlet Distribution. For example, PriorNet evidence networks \cite{ref38} utilized the subjective logic theory to build the Dirichlet distribution. In addition, ensemble distribution distillation \cite{ref39} distills from the predictions of multiple models to obtain the Dirichlet distribution. Recently, researches on multiple applications with Evidential Deep Learning (EDL) have been done, e.g., Dirichlet prior is introduced over the multinomial likelihood for evidential classification \cite{ref40,ref41,ref42}, evidential models for regression \cite{ref43,ref44}, evidential models for segmentation \cite{h1}, adversarial robustness \cite{ref45}, and calibration \cite{ref46}. Most existing methods built upon EDL are trained on evidential losses conjunct with regularization of the evidence to guide the evidence vacuity, i.e., uncertainty, behavior \cite{ref47,ref48}. In this work, we quantify uncertainty by modeling the distribution of the result with a variational Dirichlet function and minimize the boundary segmentation uncertainty by optimizing the parameters of the Dirichlet distribution.

\begin{figure*}[h]
\centering
\includegraphics[width=\linewidth]{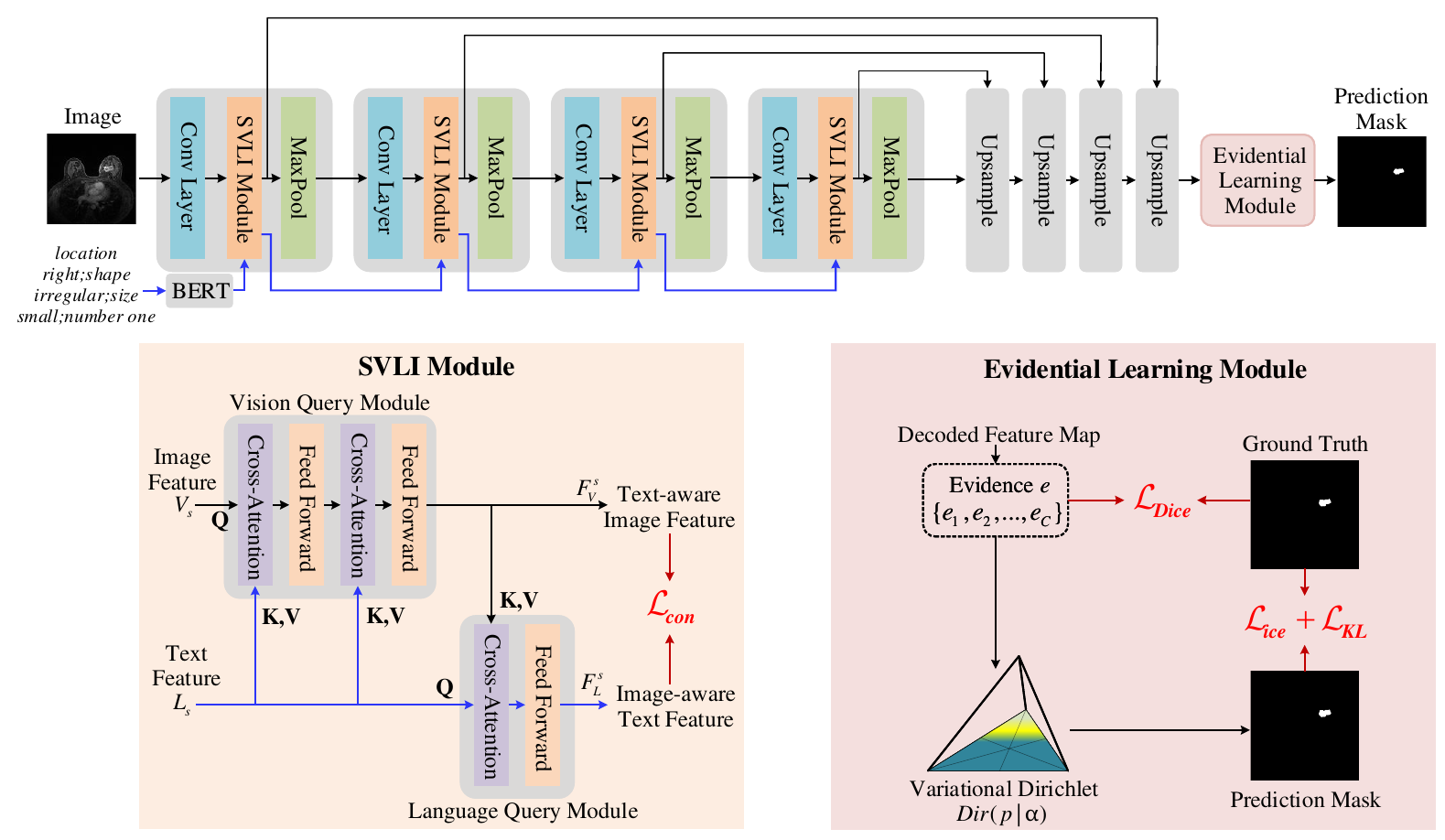}
\caption{Overview of TextBCS, consisting of the SVLI module and the evidential learning module. SVLI module performs cross fusion and alignment between the image and text feature at each downsampling stage, enhancing sufficient image-text interaction. The evidential learning module performs pixel-level uncertainty estimation based on the decoding results, preventing the model from making unreliable predictions.}
\label{fig2}
\end{figure*}

\section{Methodology}
\label{sec3}
As illustrated in Fig. \ref{fig2}, we propose the TextBCS framework that contains SVLI and EL to achieve more accurate breast cancer segmentation. The proposed framework employs DCE-MRI due to its high resolution and superior tissue contrast. In TextBCS, SVLI is designed to enables each level of image features to receive guidance from text prompts. EL is employed to estimate the uncertainty of segmentation probabilities for reliable model prediction.

\subsection{SVLI for efficient image-text interaction}
SVLI is designed with a stage-divided bidirectional cross-attention mechanism and a stage-divided cross-modal alignment loss. It performs cross-aware fusion at each down-sampling stage in the image encoder to enhance the semantic association between image and text embeddings. Given an image-text pair, the vision stage (i.e., the image encoder) takes an image as an input and the language stage takes a text prompt as an input. Each stage is indexed as $s=1,2,\cdots,S$.

\textit{(1) Stage-divided bidirectional cross-attention mechanism:} The bidirectional cross-attention mechanism alternately integrates visual and textual features of each stage, thereby extracting robust multimodal representations that mutually incorporate complementary information from the dual-modality perspectives. The visual features' contextual understanding is enhanced through cross-modal alignment with linguistically relevant information corresponding to each spatial location in the feature maps. As shown in Fig. \ref{fig2}, the vision query module of the $s$-th stage takes the text feature $L_s$ and the vision feature $V_s$ as inputs. Such vision query module conducts multi-head cross-attention by treating visual features as queries while utilizing text features as both keys and values, thus obtaining the text-aware vision feature $F_{V}^{s}\in\mathbb{R}^{H_{s}W_{s}\times C_{s}}$. The Implementation process of the vision query module is described as follows:
\begin{equation}
MCA(Q,K,V)=\mathrm{Softmax}(Q\cdot K^T/\sqrt{d_k})\cdot V
\end{equation}
\begin{equation}
f_{V}^s=FF(MCA(V_s,L_s)) + V_s
\end{equation}
\begin{equation}
F_{V}^s=FF(MCA(f_{V}^s,L_s)) + f_{V}^s
\end{equation}
where $Q,K,V$ and $d_k$ are query, key, value and the dimension of key. $MCA(\cdot)$ and $FF(\cdot)$ are the multi-head cross attention and the feed-forward. $f_{V}^s$ is the intermediate vision feature. The extracted $F_{V}^s$ is fed simultaneously into the next vision stage and the language query module.

Subsequently, the language query module takes the text-aware vision feature $F_{V}^{s}$ and the text feature $L_s$ as inputs. In language query module, the multi-head cross attention mechanism is applied. It utilizes the text feature $L_s$ as query and the text-aware vision feature $F_{V}^{s}$ as both key and value to interpret textual meanings from a visual context. Finally, the vision-aware text feature $F_L^s\in\mathbb{R}^{T\times D}$ is obtained. The Implementation process of the language query module is described as follows:
\begin{equation}
f_L^s=MCA(L_s,F_{V}^{s})
\end{equation}
\begin{equation}
F_L^s=FF(f_L^s) + f_L^s
\end{equation}
where $f_L^s$ is the intermediate text feature. The $F_L^s$ is fed simultaneously into the next language stage and the next vision query module.

\textit{(2) Stage-divided cross-modal alignment loss:} Most previous methods \cite{ref30,ref49,ref50} uses the final image-text features for cross-modal alignment. On the contrary, the developed stage-divided cross-modal alignment loss ensures more effective projection of intermediate image-text representations into a shared embedding space by aligning both low-level and high-level features across modalities.

The text-to-image contrastive loss is used for cross-modal alignment during each feature-level. As presented in Fig. \ref{fig2}, the text-aware vision feature $F_{V}^{s}$ and the vision-aware text feature $F_L^s$ are utilized to calculate the text-to-image contrastive loss $\mathcal{L}_{con}$. Before conducting cross-modal alignment, we leverage vision and language linear projections to transform both $F_{V}^{s}$ and $F_L^s$ as the same feature dimension. The text-to-image contrastive loss is calculated as follows:
\begin{equation}
\left.\mathcal{L}_{con}^{sj}=\left\{\begin{array}{ll}-\log(\sigma(\mathrm{Sim}(F_{V}^{s,j},F_L^{s,j})/\tau_s))&j\in Z^+\\-\log(1-\sigma(\mathrm{Sim}(F_{V}^{s,j},F_L^{s,j})/\tau_s))&j\in Z^-,\end{array}\right.\right.
\end{equation}
\begin{equation}
\mathcal{L}_{con}=\frac{1}{|Z|}\sum_{s=1}^S\sum_{j\in Z}\mathcal{L}_{align}^{sj}
\end{equation}
where $\mathrm{Sim}(\cdot)$ represents the calculation of cosine similarity. $\sigma$ is the sigmoid function and $\tau$ is a learnable temperature parameter. $Z,Z^+$ and $Z^-$ are the set of pixels, positive pixels and negative pixels.

\subsection{EL for uncertainty estimation of prediction probabilities}
Current medical image segmentation methods typically use Softmax to convert decoder outputs into predicted probabilities. However, Softmax tends to produce overconfident predictions \cite{ref51,ref52}, while the inherent boundary ambiguity in medical imaging introduces significant uncertainty. 

To alleviate this problem, we insert an Softplus activation function after the decoder to ensure non-negative values. These non-negative values are regarded as the evidences $e=[e_1,e_2,...,e_C]$, where $C$ is the number of classes. According to subjective logic, the Dirichlet distribution of class probabilities $Dir(p|\alpha)$ is determined from the evidences. $Dir(p|\alpha)$ provides a predictive distribution for
the segmentation results, defned as follows:
\begin{equation}
D(p\mid\alpha)=\left\{\begin{array}{ll}\frac{1}{\beta(\alpha)}\prod_{c=1}^Cp_c^{\alpha_c-1}&for \, p\in\mathcal{Q}_C\\0&otherwise\end{array}\right.
\end{equation}
where $\alpha=[\alpha_1,\alpha_2,\ldots,\alpha_C]$ is the Dirichlet distribution parameters, which is derived through $\alpha=e+1$. $p=[p_1,p_2,\ldots,p_C]$ is the probabilities that the instance is assigned to $c$-th class. $\beta(\alpha)$ is the C-dimensional multinomial beta function. $\mathcal{Q}_C$ is the $C$-dimensional unit simplex, defined as:
\begin{equation}
\mathcal{Q}_C=\left\{p|\sum_{c=1}^Cp_c=1\mathrm{~}and\mathrm{~}0\leq p_1,...,p_c\leq1\right\}
\end{equation}

Subsequently, we focus on quantifying uncertainty and optimizing parameters to minimize segmentation uncertainty. For $C$-class segmentation task, the subjective logic assigns belief mass $\{b_c\}_{c=1}^C$ to each class and an overall uncertainty mass $u$ to the whole classes frame based on the Dirichlet distribution. The $C+1$ mass values are all non-negative and their sum is one:
\begin{equation}
u_{i,j}^c+\sum_{c=1}^Cb_{i,j}^c=1
\end{equation}
where $b_{i,j}^c$ and $u_{i,j}^c$ denote the segmentation probability of
the pixel $(i,j)$ for the $c$-th class and the overall uncertainty
of the pixel $(i,j)$, respectively. Therefore, the mass of belief and uncertainty for the pixel $(i,j)$ are:
\begin{equation}
b_{i,j}^c=\frac{e_{i,j}^c}{W}=\frac{\alpha_{i,j}^c-1}{W},u_{i,j}=\frac{C}{W}
\end{equation}
where $W=\sum_{c=1}^C\left(e_{i,j}^c+1\right)=\sum_{c=1}^C\alpha_{i,j}^c$ is the Dirichlet strength. It denoets that the more evidence obtained for the $c$-th class of pixel $(i,j)$, the higher the assigned belief mass. Correspondingly, the less the total evidence obtained, the higher the overall uncertainty for the pixel $(i,j)$.

Theoretically, the optimal Dirichlet distribution on the simplex should peak sharply at the vertex representing the ground-truth class label. Consequently, its parameters must approach 1 for the correct class and remain small elsewhere. To achieve this, a loss function should be designed to directly optimize the Dirichlet parameters, minimizing segmentation uncertainty and improving boundary precision through evidence accumulation. Therefore, the cross-entropy loss function $\mathcal{L}_{ce}=\sum_{c=1}^C-y^c\log{(p^c)}$ is utilized. From the decoder's evidence, we obtain both class probabilities and the uncertainty for each pixel. Finally, the cross-entropy loss can be reformulated as:
\begin{equation}
\begin{aligned}
\mathcal{L}_{ice} & \begin{aligned} & =\int\left[\sum_{c=1}^C-y^c\log(p^c)\right]\frac{1}{\beta(\alpha)}\prod_{c=1}^C\left(p^c\right)^{\alpha^c-1}d\mathbf{p} \end{aligned} \\ & =\sum_{c=1}^Cy^c\left(\psi\left(\mathcal{Q}_C\right)-\psi\left(\alpha^c\right)\right)
\end{aligned}
\end{equation}
where $\psi\left(\cdot\right)$ is the digamma function and $y^c$ is the one-hot label. The above loss function does not guarantee that the evidence generated by the incorrect labels is lower. To address this issue, an additional term is introduced, namely the Kullback-Leibler (KL) divergence:
\begin{equation}
\begin{aligned}\mathcal{L}_{KL}&=\log\left(\frac{\Gamma\left(\sum_{c=1}^C\tilde{\alpha}^c\right)}{\Gamma(C)\sum_{c=1}^C\Gamma\left(\tilde{\alpha}^c\right)}\right)\\&\begin{aligned}+\sum_{c=1}^{C}\left(\tilde{\alpha}^{c}-1\right)\left[\psi\left(\tilde{\alpha}^{c}\right)-\psi\left(\sum_{c=1}^{C}\tilde{\alpha}^{c}\right)\right],\end{aligned}\end{aligned}
\end{equation}
where $\Gamma\left(\cdot\right)$ is the gamma function, $\tilde{\alpha}^c=y^c+(1-y^c)\odot\alpha^c$.

\subsection{Loss function}
The total loss $\mathcal{L}_{total}$ in this work is consist of segmentation loss $\mathcal{L}_{Dice}$, $\mathcal{L}_{ice}$, and $\mathcal{L}_{KL}$, defined as:
\begin{equation}
\mathcal{L}_{total}=\mathcal{L}_{Dice}+\lambda_{1}\mathcal{L}_{ice}+\lambda_{2}\mathcal{L}_{KL}+\lambda_{3}\mathcal{L}_{con}
\end{equation}
where $\lambda_{1}$, $\lambda_{2}$ and $\lambda_{3}$ are three loss balancing coefficients. The Dice loss $\mathcal{L}_{Dice}$ is used to maximize the correspondence between the segmentation prediction and the ground-truth
label, defined as:
\begin{equation}
\mathcal{L}_{Dice}=1-\sum_{i=1}^N\sum_{c=1}^C\frac{1}{NC}\cdot\frac{2|\hat{y}_{ic}\cap y_{ic}|}{(|\hat{y}_{ic}|+|y_{ic}|)}
\end{equation}
where $N$ is the number of pixels in an image and $C$ is the number of classes. $\hat{y}_{ic}$ and $y_{ic}$ are the predicted segmentation output and the ground truth, respectively.

\section{Experiments}
\label{sec4}
\begin{table*}[htbp]
  \centering
  \caption{The quantitative comparison between our method and other comparison methods on the Duke-Breast-Cancer-MRI dataset demonstrates the superiority of our method. The best values are in bold.}
  \renewcommand{\arraystretch}{1.2}
    \begin{tabular}{l|c|cc|cc}
    \hline
    Methods & Text  & Dice (\%)  & mIoU (\%) & Param (M) & Flops (G) \\
    \hline
    UNet \cite{ref54} & \ding{55}     & 81.54 & 73.22 & 14.8  & 50.3 \\
    UNet++ \cite{ref56} & \ding{55}     & 81.23 & 72.69 & 74.5  & 94.6 \\
    nnUNet \cite{ref57} & \ding{55}     & 80.15 & 71.81 & 19.1  & 412.7 \\
    TransUNet \cite{ref58} & \ding{55}     & 83.14 & 75.49 & 105   & 56.7 \\
    SwinUNet \cite{ref59} & \ding{55}     & 81.70  & 70.25 & 82.3  & 67.3 \\
    UCTransNet \cite{ref60} & \ding{55}     & 78.07 & 69.41 & 65.6  & 63.2 \\
    \hline
    CLIP \cite{ref30} & \checkmark     & 83.66 & 72.63 & 87    & 105.3 \\
    TGANet \cite{ref34} & \checkmark     & 82.11 & 71.94 & 19.8  & 41.9 \\
    ConVIRT \cite{ref32} & \checkmark     & 81.27 & 73.32 & 35.2  & 44.6 \\
    GLoRIA \cite{ref31} & \checkmark     & 82.05 & 70.57 & 45.6  & 60.8 \\
    MGCA \cite{ref61} & \checkmark     & 84.28 & 75.44 & 135.6 & 18.1 \\
    LViT \cite{ref35} & \checkmark     & 82.79 & 73.21 & 29.7  & 54.1 \\
    \hline
    Ours  & \checkmark     & \textbf{85.33} & \textbf{76.08} & 32.5  & 52.3 \\
    \hline
    \end{tabular}
  \label{table1}
\end{table*}

\begin{table}[htbp]
  \centering
  \caption{Ablation studies on the Duke-Breast-Cancer-MRI dataset demonstrate the significant improvements of the proposed innovations, including SVLI and EL. The best values are in bold.}
  \renewcommand{\arraystretch}{1.2}
    \begin{tabular}{ccc|cc}
    \hline
    Baseline & SVLI  & EL    & Dice (\%)  & mIoU (\%) \\
    \hline
    \checkmark     &      &      & 81.54 & 73.22 \\
    \checkmark     & \checkmark     &      & 84.41 & 75.24 \\
    \checkmark     &      & \checkmark     & 83.19 & 74.74 \\
    \hline
    \checkmark     & \checkmark     & \checkmark     & \textbf{85.33} & \textbf{76.08} \\
    \hline
    \end{tabular}
  \label{table2}
\end{table}

\begin{figure*}[h]
\centering
\includegraphics[width=\linewidth]{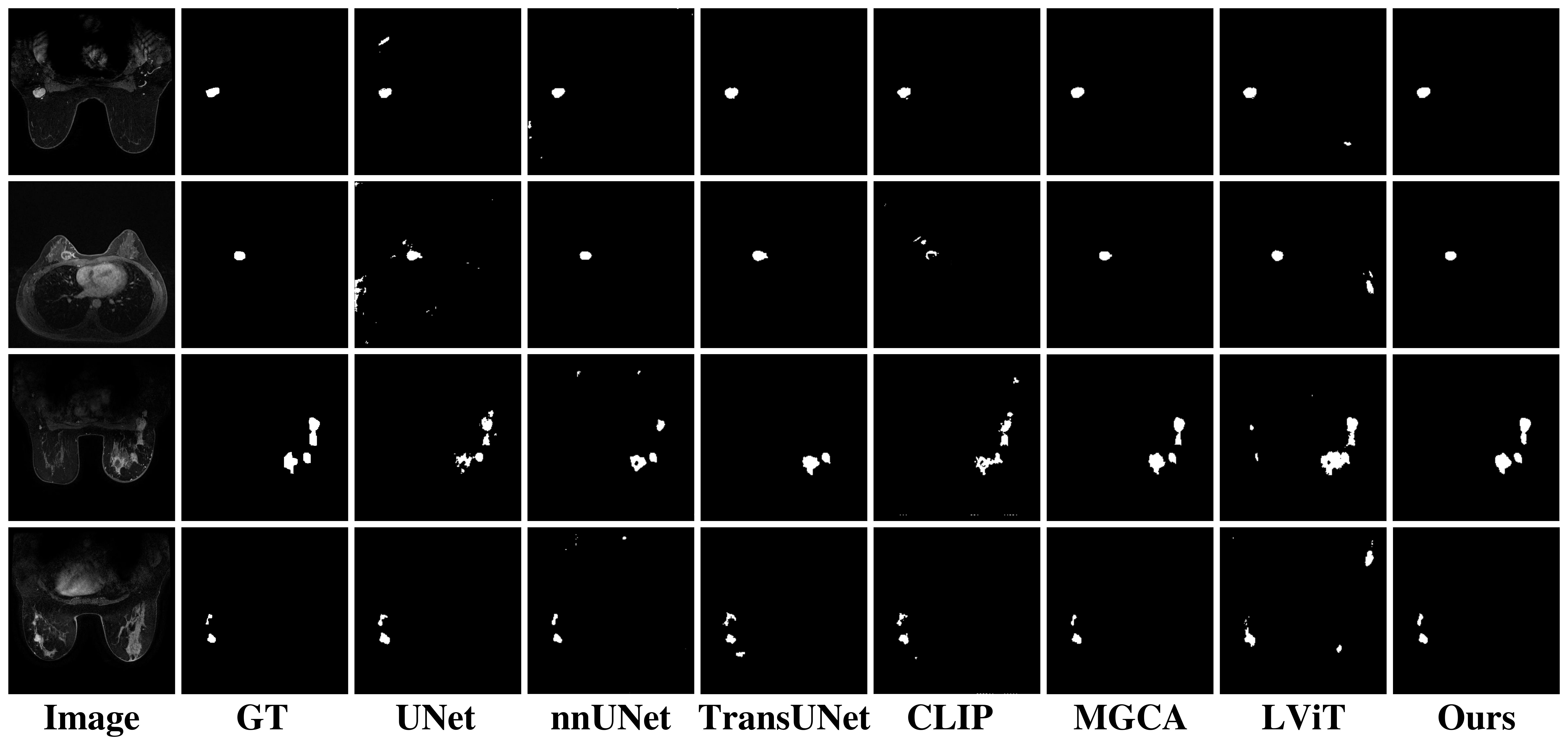}
\caption{The visual comparison of segmentation results between different methods,such as UNet, nnUNet, TransUNet, CLIP, MGCA and LViT, is displayed. Each row corresponds to one subject.}
\label{fig3}
\end{figure*}

\begin{figure*}[h]
\centering
\includegraphics[width=\linewidth]{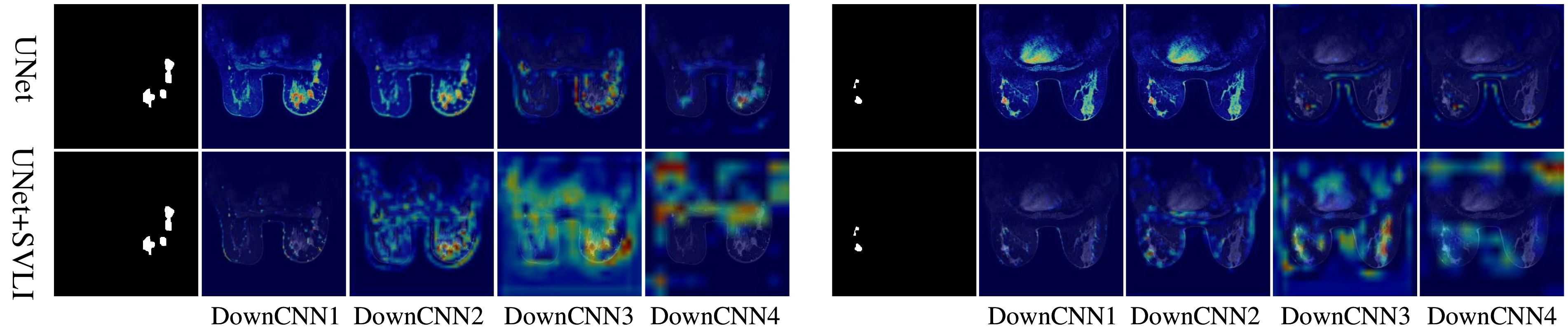}
\caption{Saliency map for interpretability study of different layers in encoder. It is obvious that the introduction of interaction between text prompts and images in each stage can better locate the breast cancer region.}
\label{fig4}
\end{figure*}

\subsection{Datasets}
We evaluate the performance of our proposed method and compare it with other UNet-based and text-guided segmentation methods on the Duke-Breast-Cancer-MRI dataset \cite{ref53}. It contains 922 patients gathered in Duke Hospital from 1 January, 2000 to 23 March, 2014 with invasive breast cancer. All MRIs were acquired in axial plane by using a 1.5 T or 3.0 T scanner (GE Healthcare and Siemens). The image size ranged from 320×320×144 to 512×512×200 with the resolution from 1.1×1.1×1.2 $mm^3$ to 0.6×0.6×1.0 $mm^3$. We extract slices containing cancer from each DCE-MRI and obtained a total of 3876 slices. These slices are divided into training set, validation set, and testing set at the patient level in a ratio of 7:1:2. For the text prompt information, two professional radiologists provide text prompts related to the cancerous area for each slice. Each text prompt was generated based on the lesion’s annotated attributes, including location, shape, size, and number. We constructed a vocabulary consisting of common descriptors (e.g., left/right, round/ ellipse/irregular, small/medium/large, one/two/three/…). Text prompts followed a consistent pattern, combining multiple attributes in short phrases. The structure format was: “location [left or right]; shape [round or ellipse or irregular]; size [small or medium or large]; number [one or two or three or …].” For example, “location left; shape irregular; size medium; number two.”

\subsection{Evaluation Metrics} 
The Dice $Dice=\sum_{i=1}^N\sum_{c=1}^C\frac{1}{NC}\cdot\frac{2|\hat{y}_{ic}\cap {y}_{ic}|}{(|\hat{y}_{ic}|+|{y}_{ic}|)}$ and mIoU $mIoU=\sum_{i=1}^N\sum_{c=1}^C\frac{1}{NC}\cdot\frac{|\hat{y}_{ic}\cap {y}_{ic}|}{|\hat{y}_{ic}\cup {y}_{ic}|}$ are used to evaluate our method and other compared methods, where $C$ is the number of categories, and $N$ is the number of pixels.

\subsection{Implementation Details}
The proposed method is implemented by Pytorch. The operating system is Ubuntu 20.04.4 LTS with 24GB RTX 3090 GPU. The UNet \cite{ref54} is used as the baseline for this work and BioClinicalBERT \cite{ref55} is utilized for encoding the text prompts to text embeddings. The Adam optimizer is utilized for model training. The batch size is 4 and the initial learning rate is $10^{-4}$. A reducing factor 0.1 is performed on the learning rate (lr = lr × 0.1) when the monitored quantity has stopped improving in five epochs. In addition, the early stop mechanism is used until the performance of model does not improve for 50 epochs, and the best trained model is used for evaluation. We set $\lambda_{1}=10^{-3}$, $\lambda_{2}=5e-7\cdot min\{1,n_{epoch}/100\}$, and $\lambda_{3}=10^{-3}$.

For baseline methods, they are reproduced using official codebases and are re-implemented using the models themselves. For each method, we perform grid search over the key hyperparameters (e.g., learning rate, batch size) within the range of the hyperparameter values provided in the original paper of each baseline method and select the best-performing configuration.

\subsection{Comparison With State-of-the-Art Methods}
\textit{(1) Comparison With UNet-based Segmentation Methods:} In this study, we conduct a thorough comparison between our proposed TextBCS framework and several UNet variant method for breast cancer segmentation. Specifically, we evaluate our method against UNet \cite{ref55}, UNet++ \cite{ref56}, nnUNet \cite{ref57}, TransUNet \cite{ref58}, SwinUNet \cite{ref59} and UCTransNet \cite{ref60}. According to the results presented in Table \ref{table1}, it is evident that the proposed TextBCS outperforms all these UNet based methods. Specifically, our TextBCS achieves a 2.19\% increase in Dice when compared to the suboptimal TransUNet. This improvement can be attributed to the advantage of our method's text prompts, which guide the model to capture the target segmentation area by utilizing text descriptions related to cancerous regions.

\textit{(2) Comparison With Text-guided Segmentation Methods:} To further evaluate the effectiveness of our TextBCS, we compare it with widely-used vision language models, including CLIP \cite{ref30}, TGANet \cite{ref34}, ConVIRT \cite{ref32}, GLoRIA \cite{ref31}, MGCA \cite{ref61} and LViT \cite{ref35}. The comparative results, as shown in Table \ref{table1}, presents the outperformance of our method in comparison to the competing methods. Our TextBCS achieves improvement of 1.05\% for the Dice metric compared to suboptimal MGCA. Although MGCA utilizes text prompts to guide segmentation, it lacks sufficient interaction between image and text and does not evaluate the reliability of segmentation results.

\textit{(3) Parameters and Computational Efficiency:} We also compared the parameters and computational efficiency of different methods. Table \ref{table1} shows the parameters (Param) and floating point operations (Flops) of each image of different models. Compared to UNet, although introducing text prompts can improve the segmentation accuracy, text-guided methods typically have more parameters and Flops due to the additional text information processing. Although text-guided methods have higher computational costs and Flops compared to some U-shaped networks like UNet and UCTransNet, they show significant performance improvements. Additionally, compared to other text-guided methods, our TextBCS has fewer Param and Flops.

\textit{(4) Qualitative comparative experimental results:} To further demonstrate the superiority of the proposed TextBCS, we provide visualization results in Fig. \ref{fig3}. In particular, each column in the figure displays the original DCE-MRI and its corresponding segmentation results from different methods.  It can be seen that our TextBCS outperforms other methods, as it can produce segmentation results more consistent with the ground truth segmentations. Moreover, TextBCS exhibits better performance in the samples with multiple cancer areas. In summary, when compared to all other methods, our method generates less false positive and negative predictions.

\begin{figure}[h]
\centering
\includegraphics[width=\linewidth]{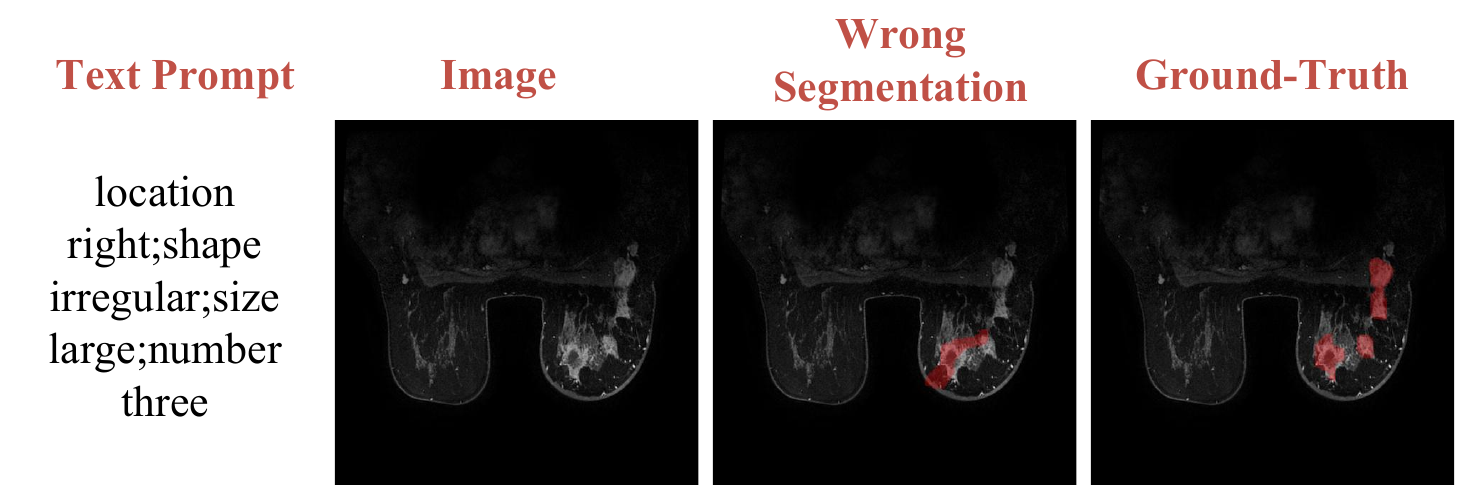}
\caption{A case that our method fail under the text prompt by segmenting incorrect areas and insufficiently identifying the target area.}
\label{fig5}
\end{figure}

\subsection{Ablation Studies}
In order to further understand the contributions of each component in our proposed TextBCS framework, we perform a series of ablation studies, with the results summarized in Table \ref{table2}. We adopt the UNet \cite{ref55} as the baseline method. The SVLI module successfully promotes mutual understanding between images and text and leads to a considerable improvement in segmentation performance, with a 2.87\% increase in Dice. Moreover, the EL can guide the model to make reliable decisions by evaluating the pixel-level uncertainty of segmentation results, resulting in performance improvement such as 1.65\% in Dice. Combining the SVLI and EL modules further boosts the network’s segmentation performance,  resulting in an increase in Dice by 3.79\%. The above experimental results show contribution of each component of the TextBCS framework in improving segmentation performance.

\begin{table}[t]
  \centering
  \caption{Ablation study of parameters ($\lambda_{1}, \lambda_{3}$). Dice values of Duke-Breast-Cancer-MRI datasets are reported. Bold values are the Dice values corresponding to the optimal parameters.}
  \renewcommand{\arraystretch}{1.2}
    \begin{tabular}{c|cccc}
    \hline
    Coefficient    & 1.0     & 0.1   & 0.01  & 0.001 \\
    \hline
    $\lambda_{1}$    & 84.79 & 84.83 & \textbf{85.33} & 85.14 \\
    $\lambda_{3}$    & 85.11 & 84.90 & \textbf{85.33} & 85.07 \\
    \hline
    \end{tabular}
  \label{table3}
\end{table}

\begin{table}[t]
  \centering
  \caption{P-values from t-test between our method and other methods shown that our method exhibits statistical significance. P-values are calculated based on the Dice.}
  \renewcommand{\arraystretch}{1.2}
    \begin{tabular}{lc}
    \hline
    Methods & p-values \\
    \hline
    Unet \cite{ref54}  & 0.0014 \\
    Unet++ \cite{ref56} & 0.0032 \\
    nnUNet \cite{ref57} & 0.0008 \\
    TransUNet \cite{ref58} & 0.0030 \\
    SwinUNet \cite{ref59} & 0.0018 \\
    UCTransNet \cite{ref60} & 0.0009 \\
    CLIP \cite{ref30} & 0.0137 \\
    TGANet \cite{ref34} & 0.0074 \\
    ConVIRT \cite{ref32} & 0.0106 \\
    GLoRIA \cite{ref31} & 0.0082 \\
    MGCA  \cite{ref61} & 0.0316 \\
    LViT  \cite{ref35} & 0.0065 \\
    \hline
    \end{tabular}
  \label{table4}
\end{table}

\subsection{Parameter Analysis}
In the optimization of our method, the predefined parameter $\lambda_{1}$ and $\lambda_{3}$ is used to balance the weights of the evidential learning loss and the cross-modal alignment loss. The Dice values with different $\lambda_{1}$ and $\lambda_{3}$ are displayed in Table \ref{table3}, in which the values of $\lambda_{1}$ and $\lambda_{3}$ ranges from 0.001 to 1.0. It is observed that the parameters $\lambda_{1}=0.01$ and $\lambda_{3}=0.01$ achieve the best segmentation performance as 85.33\%.  It has also been noted that increasing or decreasing $\lambda_{1}$ and $\lambda_{3}$ not necessarily increases the segmentation performance. The possible reason is that the evidential learning module dominates the optimization with too large $\lambda_{1}$ and thus more attention is paid to make the gradient converge toward the direction that minimizes the uncertainty of the segmentation results rather than the segmentation.

\begin{figure}[h]
\centering
\includegraphics[width=\linewidth]{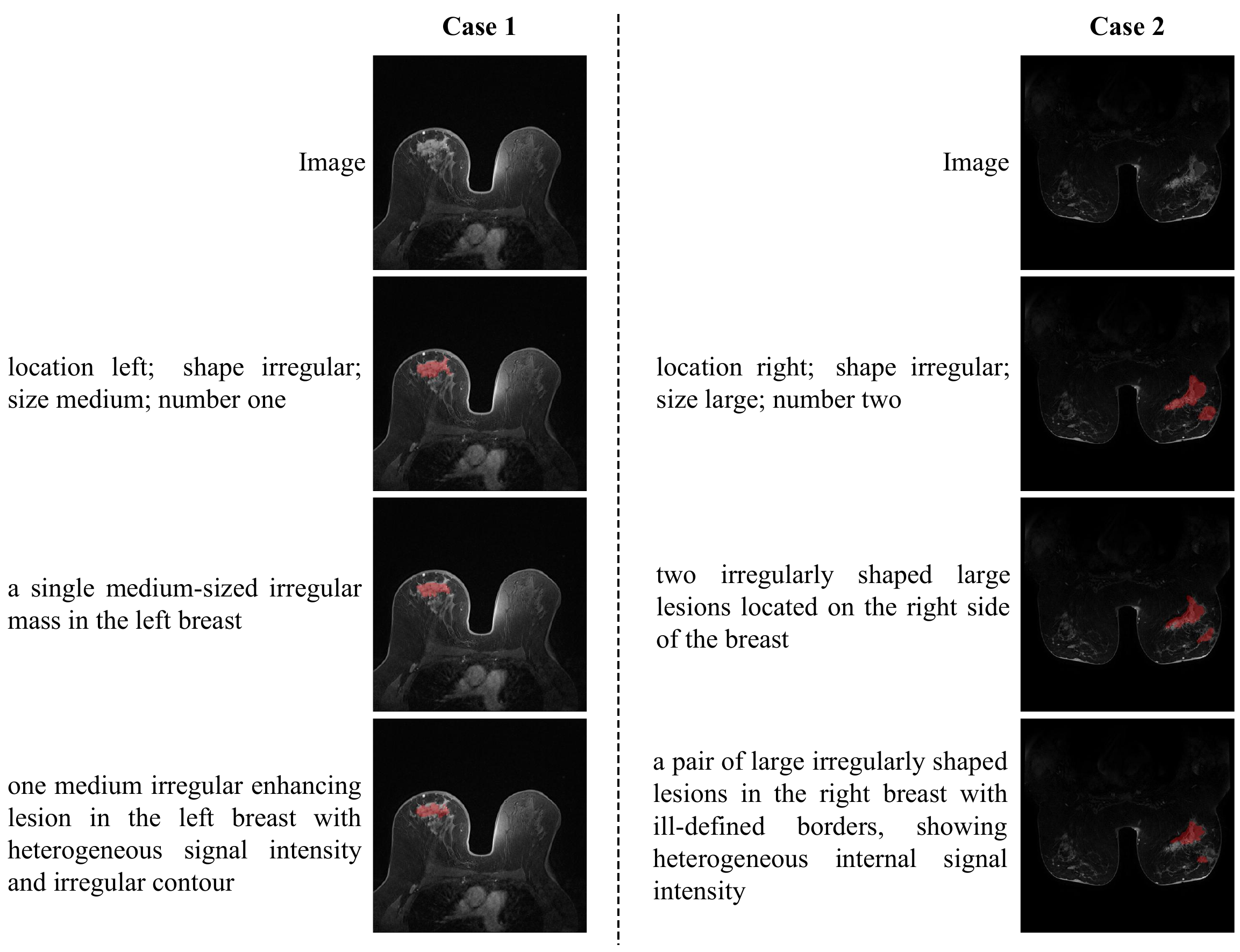}
\caption{Qualitative experiment of different styles of text prompts shows the robustness of the model to variations in the text prompts.}
\label{fig6}
\end{figure}

\subsection{Statistical Analysis}
To verify whether the proposed method has statistical significance compared to other methods, the paired t-test with $a$ = 0.05 is conducted between our method and other methods, and the p-values are reported based on Dice. As shown in Table \ref{table4}, the p-values are all less than 0.05 under a 95\% confidence. Therefore, based on the above analysis, it can be concluded that the proposed method is statistically significant.

\subsection{Interpretability Study}
The interpretability study is performed to explore whether the TextBCS can notice cancer regions and whether the introduction of text information can enhance the attention to cancer regions. As shown in Fig. \ref{fig4}, we conduct experiments on two cases. We perform activation mapping in four down-sampling stages, i.e., DownCNN1, DownCNN2, DownCNN3 and DownCNN4. The text information is input to the model through SVLI module, thus the difference in activation regions between UNet and UNet+SVLI can be approximated as the difference brought by the text prompt. It can be seen that the activation effect of the region of interest of UNet+SVLI is similar to that of Ground Truth. It is also worth noting that image feature extracted by DownCNN3 and DownCNN4 of UNet fails to activate the cancer regions. However, the image feature extracted by UNet+SVLI can directly activate the relevant cancer region by introducing the text information. It indicates that the text information can effectively help locate cancer region in the breast, thus prompting the network to pay more attentions on the region indicated by the text information.

\subsection{Robustness Analysis to Different Styles of Text Prompts}
To assess the model’s robustness to different styles of text prompts, we conducted a qualitative analysis using diverse text prompts describing the same lesion with different styles. As shown in Fig. \ref{fig6}, for the text prompt of case 1, “location left; shape irregular; size medium; number one”, we used ChatGPT 5 to generate its variants, such as "a single medium-sized irregular mass in the left breast" and "one medium irregular enhancing lesion in the left breast with heterogeneous signal intensity and irregular contour". Similarly, for case 2, “location right; shape irregular; size large; number two”, we also generated two diverse variants, i.e., "two irregularly shaped large lesions located on the right side of the breast" and "a pair of large irregularly shaped lesions in the right breast with ill-defined borders, showing heterogeneous internal signal intensity". The qualitative experiment in Fig. \ref{fig6} showed that this model exhibited stable segmentation performance in diverse variants of text prompts. However, when the text prompts are complex, as shown in the bottom of Figure 5, there are some differences between the segmentation results of complex text and the original text. This is because the complex text introduces some vocabularies that the model did not learn during the training process (because the model was trained on the original text). Overall, when the text prompt varies, the segmentation results would not differ too much, proving the robustness of the model. In future work, diverse styles of text prompts can be introduced to training to further enhance the model's generalization ability.

\subsection{Failure Case Analysis}
While our TextBCS has already demonstrated promising results, there remains much to explore and improve, particularly in text prompts. For instance, in certain scenarios, wrong segmentation occurs due to insufficient text prompt, as shown in Fig. \ref{fig5}. For the text prompt “location right; shape irregular; size large; number three” showed in Fig. \ref{fig5}, “right” does not clearly specify which quadrant of the breast is referred to; “irregular” lacks morphological context (e.g., irregular contour or lobulated margin); and “large” has no explicit reference scale. Consequently, the model may struggle to achieve precise segmentation using such ambiguous text prompt. To address 
this issue, future enhancements could involve adapting the model to accept a broader array of prompt types, such as points and boxes, adopting more fine-grained text prompts.

\subsection{Practical Application}
\noindent
\textbf{Text Prompts.} In our current study, the text prompts were provided by two experienced radiologists. However, we agree that relying on manual prompt generation would limit scalability in real-world clinical applications. In practice, the proposed method does not require radiologists to write text prompts for each scan. Instead, there are two deployment strategies.

(1) Employ large language models (LLMs) such as ChatGPT and Gemini-2.5-Pro to automatically generate text prompts for each scan. In this paradigm, the image can be fed into a multimodal LLM that has been adapted for medical imaging, which then generates concise, standardized text prompts describing relevant lesion attributes, such as location, shape, size, and number. This strategy would transform vision information into text prompts in real time, without the need for manual annotation. Although precise text prompts generated by a powerful LLM can understand medical images well, such descriptions primarily convey coarse-grained contextual understanding rather than detailed spatial localization. In other words, even though the LLM demonstrates a high-level comprehension of lesion attributes, it lacks the pixel-level precision required for accurate segmentation. Therefore, text alone cannot delineate the lesion boundaries. In contrast, by jointly leveraging text prompts and image data during segmentation network training, it can effectively associate semantic concepts with their corresponding spatial regions, leading to precise segmentation outcomes. However, if the LLMs generate incorrect text prompts, it can negatively impact downstream segmentation tasks. Therefore, a direction for future work is to assess the reliability of generated text prompts using feature-level similarity metrics, such as cosine similarity between image embeddings and their corresponding text embeddings to filter out incorrect text prompts.

(2) Another strategy involves extracting prompts from existing radiological reports. Hospital information systems store reports written by radiologists. In practice, radiological reports can be converted into standardized text prompts following the same attribute template used during training. This strategy leverages routinely collected clinical data and minimizes additional workload for radiologists.

\noindent
\textbf{Patient Privacy.} While LLMs offer a convenient method to automatically generate text prompts from medical images, their use in clinical settings raises serious privacy concerns. Transmitting MRI scans to external servers can expose patient information. To mitigate these risks, future implementations should prioritize internal deployment or locally fine-tuned LLMs for medical image analysis.

\section{Conclusion}
In this work, we propose a text-guided breast cancer segmentation method (TextBCS), which can effectively use text descriptions related to cancer regions to locate breast cancer regions. In addition, a SVLI module is proposed to promote image-text interaction at each feature-level of down-sampling. And the EL module is introducing to evaluate the reliability of segmentation results. Experimental results on a public breast cancer segmentation dataset shows better performance of our TextBCS compared to the SOTA segmentation methods.

\section*{CRediT authorship contribution statement}
\textbf{Jingxing Zhong:} Writing – original draft, Investigation, Validation. \textbf{Qingtao Pan:} Methodology, original draft. \textbf{Xuchang Zhou:} Software, Formal analysis, Data curation. \textbf{Jiazhen Lin:} Writing – review \& editing, Supervision, Resources. \textbf{Xinguo Zhuang:} Funding acquisition, Project administration, Supervision.

\section*{Declaration of competing interest}
The authors declare that they have no known competing financial
interests or personal relationships that could have appeared to influence the work reported in this paper.

\section*{Acknowledgement}
This work is supported by the Xiamen Medical and Health Guidance Project (No. 3502Z20224ZD1014), the Natural Scientific Foundation of Xiamen (No. 3502Z20227340), the Fujian Natural Science Foundation of China (No. 2022J011372), Young Investigator Research Program of Xiang’an Hospital of Xiamen University (XM02070002n), Open Innovation Fund for undergraduate students of Xiamen University (No. KFJJ-202410).



\bibliographystyle{elsarticle-num} 
\bibliography{ref}
\end{document}